\pdfoutput=1

\documentclass[11pt]{article}

\usepackage[final]{acl}

\usepackage{times}
\usepackage{latexsym}

\usepackage[T1]{fontenc}

\usepackage[utf8]{inputenc}

\usepackage{microtype}

\usepackage{inconsolata}

\usepackage{graphicx}
\usepackage{multirow}
\usepackage{colortbl}
\usepackage{algorithm}
\usepackage{algorithmic}
\usepackage{booktabs} 
\usepackage{amsmath}
\usepackage{tabularx}
\usepackage{utfsym}
\usepackage{makecell}
\usepackage[normalem]{ulem}
\useunder{\uline}{\ul}{}
\usepackage{authblk}
\title{{\scshape MSDiagnosis}: A Benchmark for Evaluating Large Language Models in Multi-Step Clinical Diagnosis}

\author{
Ruihui Hou$^1$~~~
Shencheng Chen$^1$~~~
Yongqi Fan$^1$~~~
Guangya Yu$^1$~~~
Lifeng Zhu$^2$~~~
Jing Sun$^2$~~~ \\
Jingping Liu$^1$~~~ 
Tong Ruan$^1$~~~
\smallskip 
\\
$^1$School of Information Science and Engineering, \\
East China University of Science and Technology, Shanghai, China
\\
$^2$Ruijin Hospital, Shanghai Jiao Tong University School of Medicine, Shanghai, China
}

\begin{document}
\maketitle
\begin{abstract}

Clinical diagnosis is critical in medical practice, typically requiring a continuous and evolving process that includes primary diagnosis, differential diagnosis, and final diagnosis. 
However, most existing clinical diagnostic tasks are single-step processes, which does not align with the complex multi-step diagnostic procedures found in real-world clinical settings.
In this paper, we propose a Chinese clinical diagnostic benchmark, called MSDiagnosis. 
This benchmark consists of 2,225 cases from 12 departments, covering tasks such as primary diagnosis, differential diagnosis, and final diagnosis.
Additionally, we propose a novel and effective framework. 
This framework combines forward inference, backward inference, reflection, and refinement, enabling the large language model to self-evaluate and adjust its diagnostic results. 
To this end, we test open-source models, closed-source models, and our proposed framework.
The experimental results demonstrate the effectiveness of the proposed method.
We also provide a comprehensive experimental analysis and suggest future research directions for this task.

\end{abstract}

\section{Introduction}

\begin{figure}[ht]
\centering
\includegraphics[width=1\columnwidth]{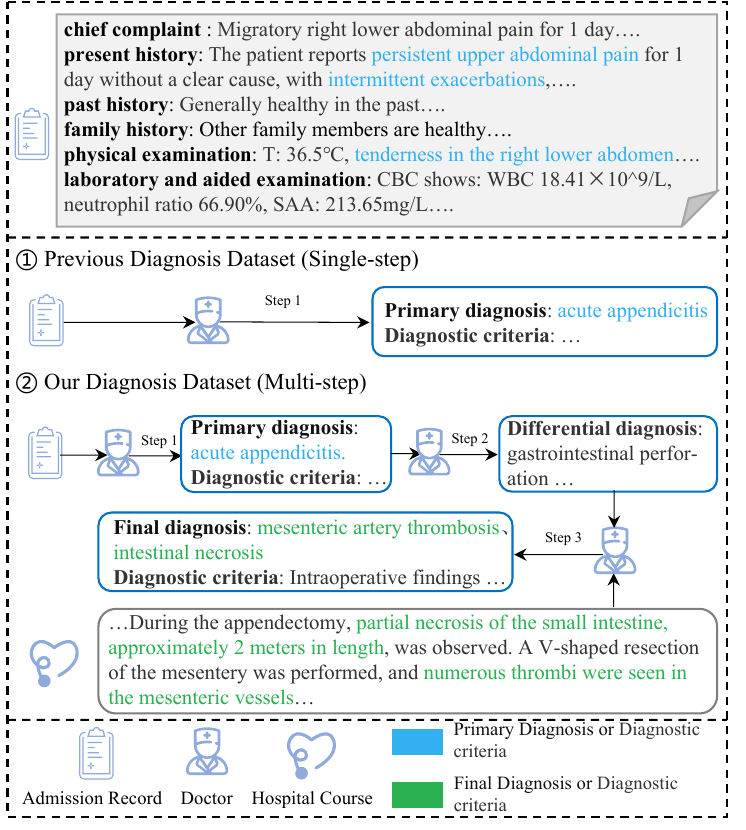} 
\caption{An example of our diagnostic benchmark and its differences from the previous diagnostic benchmark.}
\label{example}
\end{figure}



Clinical diagnosis is a central element of clinical decision-making, involving the integration of chief complaint, present history, and physical examinations, along with clinical experience and medical knowledge, to make a scientifically grounded judgment regarding the nature and cause of a disease~\cite{ball2015improving}. 
Accurate diagnosis helps to provide the most appropriate treatment for patients.
In recent years, research in data-driven clinical diagnosis has advanced rapidly, particularly with the widespread use of electronic medical records (EMRs), which enable diagnostic models to integrate multi-stage patient data for more precise analysis~\cite{time-series}.


\begin{table*}[ht]
\centering
\setlength{\tabcolsep}{0.5mm}
\small
\caption{Overview of the clinical diagnosis benchmark. ``*'' indicates that the dataset contains data for multiple tasks, with the diagnosis being just one of them. In the ``Size'' column, the value outside the brackets indicates the number of samples related to diagnosis, and the value inside the brackets indicates the total number of samples in the dataset. 
The terms ``Criteria'', ``Multi-Step'', ``Q.Type'', and ``MCQA'' stand for ``diagnostic criteria'', ``multi-step diagnosis'', ``question type'', and ``multiple-choice QA'' respectively.
}
\begin{tabular}{lcccccc}
\toprule
\textbf{Dataset}    & \textbf{Data Source}    &\textbf{Q.Type}    & \textbf{Criteria}  
  & \textbf{Multi-Step}    & \textbf{Key Points}  & \textbf{Size}        \\ \midrule
DDx-basic~\cite{medbench}         &  Examination Paper  & MCQA          & \usym{2717}             & \usym{2717}       & \usym{2717}     & 150         \\
DDx-advanced~\cite{medbench}      &  Examination Paper  & MCQA          & \usym{2717}             & \usym{2717}       & \usym{2717}    & 30          \\
CMExam*~\cite{cmexam}           & Medical Examination & MCQA          & \usym{2717}               & \usym{2717}       & \usym{2717}    & 1,630(60K+) \\
AgentClinic-MedQA~\cite{agentclinic} & GPT4+MedQA    & OpenQA       & \usym{2717}                 & \usym{2717}       & \usym{2717}    & 107         \\
AI Hospital~\cite{aihospital}      & Medical Website    & OpenQA       & \usym{1F5F8}             & \usym{2717}       & \usym{2717}    & 506      \\
CMB-Clin~\cite{cmb}          & Medical Textbook  & OpenQA       & \usym{1F5F8}                    & \usym{2717}       & \usym{2717}      & 74          \\
RJUA-QA~\cite{rjuaqa}           & Synthetic      & OpenQA       & \usym{1F5F8}                    & \usym{2717}       & \usym{2717}      & 2,132       \\
Ours              & Medical Website              & OpenQA       & \usym{1F5F8}                    & \usym{1F5F8}      & \usym{1F5F8}      & 2,225       \\ \bottomrule
\end{tabular}
\label{compare_data}
\end{table*}

The clinical diagnosis process is a continuous and evolving process~\cite{clinical-decision}. 
Specifically, doctors first make a primary diagnosis based on the patient's chief complaint, present history, and laboratory and aided examination. 
Then they narrow down the diagnostic options through differential diagnosis. Finally, they determine the final diagnosis by considering clinical changes throughout the diagnostic and treatment process.
However, a significant gap remains between the existing diagnostic benchmark and actual clinical practice. Most existing diagnostic tasks are single-step processes~\cite{cmb, rjuaqa, cmexam}, where a diagnosis is made directly based on the patient's medical history, chief complaint, and examination results. This single-step approach does not align with the multi-step process typically used in clinical practice.
In addition, current evaluations of single-step diagnostic tasks typically use BLEU and ROUGE metrics. However, these metrics, which rely solely on lexical overlap, cannot effectively assess the rationale behind the diagnosis or the coverage of key evidence.


Hence, in this paper, we propose a \textbf{M}ulti-\textbf{S}tep clinical \textbf{Diagnosis} benchmark (MSDiagnosis). 
In this benchmark, a primary diagnosis and diagnostic criteria are generated based on the patient’s chief complaint, present history, and examination results. Then, a differential diagnosis is made to narrow down the possible diseases. Finally, the final diagnosis and diagnostic criteria are made by combining the hospital course, primary diagnosis, and differential diagnosis. The specific process is shown in Fig.~\ref{example}.
The MSDiagnosis benchmark is designed to evaluate the diagnostic capabilities of large language models (LLMs). The benchmark comprises 2,225 medical records collected from medical websites, covering 12 departments. 
Each medical record contains five related diagnostic questions. 
For ease of comparison, we list the most relevant works in Table~\ref{compare_data}. 
Additionally, to accurately assess diagnostic criteria, we annotate each answer with corresponding key points and evaluate the quality of the answers by calculating the macro-recall for each key point.


For the multi-step clinical diagnostic benchmark, we propose a simple and effective pipeline framework. This framework consists of two stages. The first stage is forward inference. 
Specifically,  we retrieve similar EMRs to serve as in-context learning (ICL), allowing the LLM to diagnose the patient. 
The second stage involves backward inference, reflection, and refinement. 
Specifically, we design a backward inference strategy, a reflection strategy, and a refinement strategy, enabling the LLM to check and refine its diagnostic results.

Our contributions are summarized as follows:
\begin{itemize}
    %
    \item 
    We propose a multi-step clinical diagnosis benchmark (MSDiagnosis) that includes three tasks: primary diagnosis, differential diagnosis, and final diagnosis. 
    \item We propose a simple and efficient framework that combines forward and backward inference, enabling LLMs to validate and refine diagnostic results.
    \item 
    We evaluate both open-source and closed-source LLMs on MSDiagnosis and conduct extensive experiments on our framework. The experimental results demonstrate the effectiveness of the proposed method.

\end{itemize}

\section{Problem Formulation}



In this paper, we consider the multi-step diagnosis task as a multi-round dialogue problem. For each complex EMR, we simulate the interaction between an examiner and a candidate, where the candidate is required to provide an answer to a specific diagnostic question.
Specifically, given a patient's admission record $E$, which includes the chief complaint, present history, past history, physical examination, and laboratory and aided examination, the candidate answers two questions $R_1 = [Q_1, Q_2]$ in the first round of dialogue. 
These questions mainly involve the patient's primary diagnosis and its diagnostic criteria. 
As illustrated in Fig.~\ref{framework}, $Q_1$ and $Q_2$ are ``What is the patient’s primary diagnosis?'' and ``What is the criterion for the primary diagnosis?'' respectively.
In the second round of dialogue, the candidate needs to answer the question $R_2 = [Q_3]$. This question refers to asking the patient about their differential diagnosis. 
In the third round of dialogue, the candidate receives the patient's hospital course $T$, the dialogue history $H$, and two additional questions $R_3 = [Q_4, Q_5]$. 
These questions focus on the patient's final diagnosis and its diagnostic criteria. Therefore, the input $X$ for the current dialogue can be formally expressed as $X = E + T + H + Q_i$, where $Q_i \in R_3$. 
The reference answer for this input is $A_i$.

\section{MSDiagnosis}

In this section, we detail the construction process of the dataset, including the data collection, selection, and data annotation.

\subsection{Data Collection and Selection}


In this study, we select a Chinese medical website\footnote{\url{https://www.iiyi.com/}} as the source of EMRs. 
The raw data are anonymized, and we apply additional anonymization to ensure no protected health information (PHI) remains.
After de-identifying the data, we obtain a total of 11,900 EMRs. 
To ensure high-quality EMR, we select the data through four steps. 
First, we remove redundant information unrelated to medical content, such as web elements, copyright statements, and advertisements. 
Second, we eliminate EMRs with duplicate field values and missing fields, such as those lacking primary diagnosis, differential diagnosis, final diagnosis, or hospital course fields. 
After this step, we retain 5096 EMRs.
Third, we deduplicate EMRs from the same department. If two records have identical chief complaints, present histories, and physical examinations, one of them is removed.
After this step, we retain 4179 EMRs.
Fourth, to avoid duplicate patients in the dataset, we conduct a meticulous matching process using patients’ demographic information (such as gender, occupation, etc.) and filter out individuals with identical profiles. After rigorous screening, we ensure the dataset contains no patients with fully identical records.
After four steps, we ultimately obtain 3,501 high-quality, complex, and authentic EMRs.

\subsection{Data Annotation}
\label{annotation}
\subsubsection{Question Construction}
\label{question_construction}

We first manually construct five seed questions, including the patient's primary diagnosis and its criteria, the differential diagnosis, and the final diagnosis and its criteria. The definition of each question is shown in Appendix~\ref{question_denfinition}.
Then, we use GPT-4 to expand each constructed question with ten similar questions. 
Finally, we manually check and filter out unreasonable question expressions.

\subsubsection{Answer Annotation}
\label{answer_annotation}
To ensure the quality of the dataset construction, we form a professional team comprising three inspectors and one reviewer. All team members undergo specialized medical training to understand primary diagnosis, differential diagnosis, and final diagnosis. The construction process includes three stages: first-round annotation, second-round checking, and third-round review.

\textbf{First-round annotation}. 
For questions related to initial diagnosis, differential diagnosis, and final diagnosis, if the original medical records contain relevant answers, the standard answers are directly extracted from the EMR. 
However, statistical analysis reveals that 1,989 cases lack diagnostic results. 
Additionally, since the original data does not provide criteria for the primary and final diagnoses, we initially use GPT-4 to generate diagnostic criteria for 3,501 cases.

\textbf{Second-round checking}. 
We engage three university students to simultaneously check the rationality of all question-answer pairs. 
Samples that all three inspectors considered unreasonable are directly discarded. 
If one or two inspectors find a sample unreasonable, it is manually re-annotated and retained only if all three inspectors subsequently agree on its reasonableness.
After rigorous check, we identify 1,222 cases that are unanimously deemed unreasonable.

\textbf{Third-round review}. 
A verified batch is given to a medical expert for double review.
The medical expert randomly inspects 20\% of the batch samples. Any unqualified annotations are returned to the check team with explanations, which can further refine the standards. This process is repeated until the batch accuracy reaches 95\%. 
After this stage, we remove 54 cases, ultimately retaining 2,225 high-quality cases.

\subsubsection{Key Points Annotation}

To more accurately evaluate open-ended questions such as primary diagnostic criteria and final diagnostic criteria, we annotate the key points of the answers to these questions. Specifically,
for each primary and final diagnostic criteria, we first construct key point extraction prompts and employ GPT-4 to extract the information from the standard answers. The key point for the diagnostic criteria mainly includes four categories: medical history, symptoms, physical signs, and examination results.
Then, we invite two students with specialized medical training to validate the extracted information. 
If any inconsistencies are found in the EMRs, the data is re-annotated. 
Finally, we obtain 2,225 high-quality samples to construct MSDiagnosis.
We calculate the Cohen’s Kappa~\cite{Cohen} score for the key points of the primary diagnostic criteria and the final diagnostic criteria. The Cohen’s Kappa for the primary diagnostic criteria is 0.81, indicating a high level of agreement, while the Cohen’s Kappa for the final diagnostic criteria is 0.79, indicating a moderate level of agreement.

\section{Dataset Analysis}

In this section, we introduce the statistics and the characteristics of the MSDiagnosis in detail.

\subsection{Data Statistics}


\begin{figure}[t]
\centering
\includegraphics[width=\columnwidth]{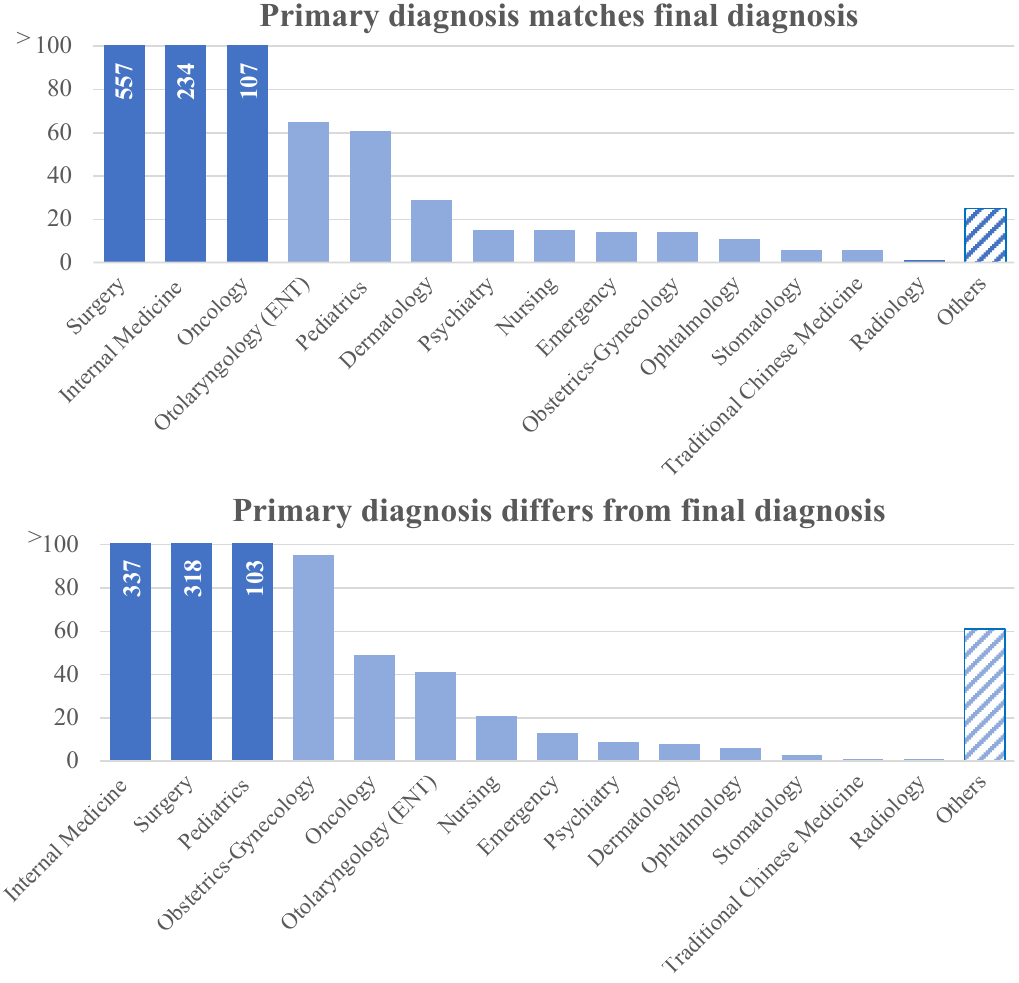} 
\caption{Distribution of ``Primary diagnosis matches final diagnosis'' and ``Primary diagnosis differs from final diagnosis'' in MSDiagnosis across different departments.}
\label{fenbu}
\end{figure}

\begin{table}[t]
\centering
\caption{The statistics of our constructed dataset. ``Diag-M'' and ``Diag-D'' refer to EMRs where the primary diagnosis is the same as or different from the final diagnosis, respectively. ``AvgP'' and ``AvgF'' represent the average number of diseases in the primary and final diagnoses, respectively.}
\setlength{\tabcolsep}{2.5mm}
\small
\begin{tabular}{lccccc}
\toprule
\textbf{Type}  & \textbf{Size} & \textbf{Diag-M} & \textbf{Diag-D} & \textbf{AvgP} & \textbf{AvgF} \\ \midrule
Train & 1,557 & 808                    & 749                    & 2.53 & 2.78 \\
Test  & 445  & 234                    & 211                    & 2.44 & 2.77 \\
Dev   & 223  & 118                    & 105                    & 2.44 & 2.88 \\
Total & 2,225 & 1,160                   & 1,065                   & 2.51 & 2.79 \\ \bottomrule
\end{tabular}
\label{table2}
\end{table}

As reported in Table~\ref{table2}, the dataset contains a total of 2,225 EMRs covering 12 departments, which are divided into training, validation, and test sets according to a 7:1:2 ratio. 
Additionally, we conduct a detailed statistical analysis of patients’ diagnoses from three main perspectives.
1) \textit{Department Distribution}.
Fig.~\ref{fenbu} presents the distribution of EMRs across different departments, categorized by whether the primary diagnosis is consistent with the final diagnosis.
In the category where the primary and final diagnoses are the same, surgery has the highest proportion (48.02\%). 
In the category where the primary and final diagnoses are different, internal medicine has the highest proportion (31.64\%).
2) \textit{Number of Diseases}. 
The average number of diseases diagnosed in the primary diagnosis stage is 2, with a maximum of 21 diseases. 
At the final diagnosis stage, the average number of diseases is 3, with a maximum of 21 diseases.
3) \textit{Types of Diagnostic Changes}. Statistical analysis reveals that diagnosis changes primarily fall into three categories: addition, deletion, and modification.

\begin{figure*}[t]
\centering
\includegraphics[width=2\columnwidth]{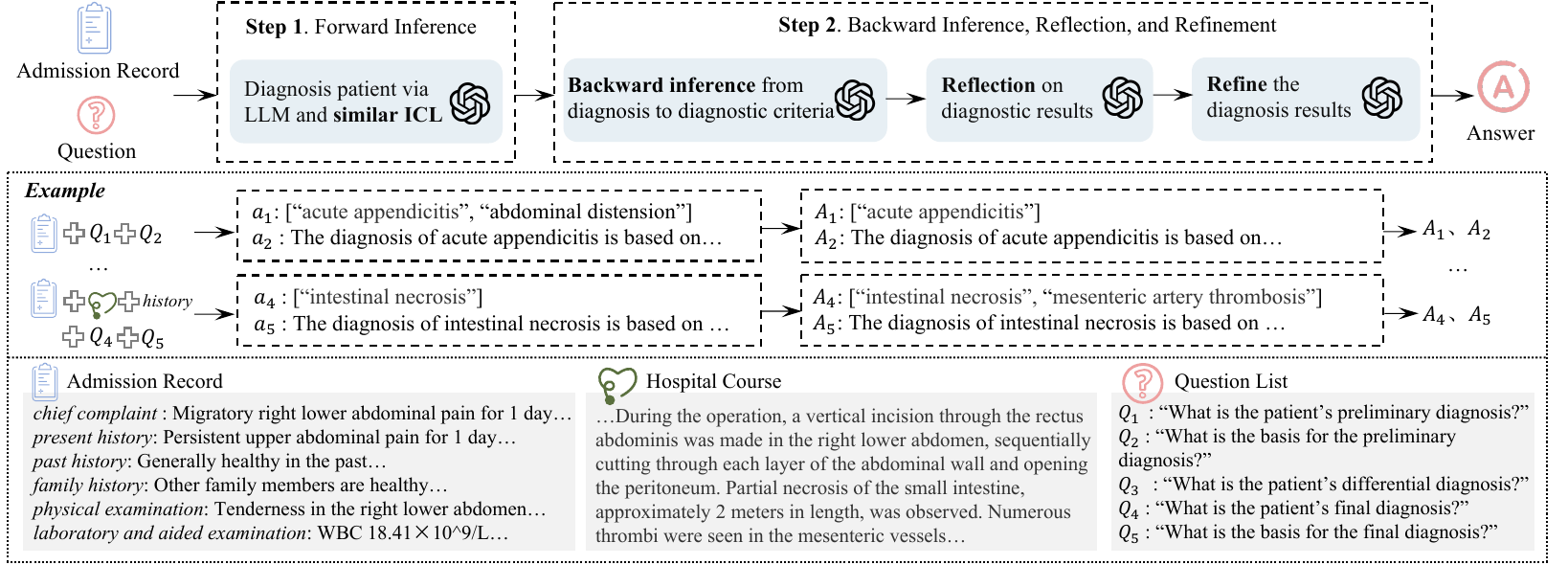} 
\caption{Our framework for the multi-step clinical diagnosis. The top portion of the figure illustrates the flow of the framework, comprising two stages. The first stage involves the forward inference diagnosis. The second stage focuses on backward inference, reflection, and refinement.}
\label{framework}
\end{figure*}

\subsection{Data Characteristics}


The MSDiagnosis has several significant advantages compared to previous medical diagnostic datasets:
1) \textit{Reliability and Authenticity of Data}. The MSDiagnosis is entirely based on EMRs, including detailed treatment plans and course records. 
2) \textit{Rich Variety of Departments and Diseases}. The MSDiagnosis covers patient EMRs from 12 different departments and includes various types of diseases, such as common diseases, rare diseases, acute diseases, and chronic diseases. This wide coverage of departments and diverse disease types ensures the broad applicability of the data.
3) \textit{Inclusion of Multi-step Diagnostic Processes}. 
This multi-step diagnostic process is more in line with actual clinical diagnostic scenarios. By recording diagnostic changes, the dataset better reflects the complex diagnostic paths and dynamic changes in real-world medical practice.

\section{Method}

In this section, we first provide an overview of our framework. Then, we describe each module of the framework in detail.

\subsection{Framework}


As illustrated in Fig.~\ref{framework}, our proposed framework mainly consists of two stages. 
The first stage involves forward inference. In this stage, we retrieve similar EMRs to serve as ICL, enabling the LLM to diagnose the patient. 
The second stage is backward inference, reflection, and refinement. 
In this stage, we first validate the diagnostic criteria against the facts derived from the diagnostic results. 
Then, the LLM uses designed reflection rules to evaluate the diagnostic outcomes. Finally, the diagnosis is refined by integrating all the previous results.

\subsection{Forward Inference}


This part aims to make diagnoses for patients based on admission records. In this paper, we utilize an LLM (e.g., GPT4o-mini) with the ICL method to achieve this purpose. Its core idea is to select similar EMRs from the training set, guiding the model in making accurate diagnoses for patients.

Specifically, given the admission record $E$ and question $Q_o$, where $o$ is the total number of questions, we select ICL examples with similar semantics to $E$ through the following steps. 
First, we utilize the BGE~\cite{bge_embedding} model to obtain the representations of $E$ and each sample $Y_i\ (1\leq i\leq u)$ in the training set, where $u$ is the total number of the training set.
Second, we calculate their cosine similarity and select the top $K$ samples with the highest similarity as ICL examples.
In addition to using $E$ and the top $K$ samples as input to the LLM, we introduce a role definition and a predefined rule.
The role definition is phrased as ``You are a professional doctor, and you need to complete diagnosis task''. 
The rules for constraining the output format of the LLM refer to ``The output format of the diagnostic results can be loaded directly using the JSON.load() function''.
Finally, the LLM would generate the answer $a_o$ corresponding to question $Q_o$.
The detailed prompt is shown in Appendix~\ref{prompt}.

\begin{table*}[ht]
\small
\centering
\caption{Method comparisons on MSDiagnosis (\%). 
``Pri'' refers to the primary diagnosis. ``Fin'' stands for the final diagnosis. ``DD'' means differential diagnosis. 
All results are the average of three runs. 
The best and second results, excluding the human method, are highlighted in \textbf{bold} and {\ul underline}, respectively.}
\setlength{\tabcolsep}{1.3mm}
\begin{tabular}{lcccccccccccccccc}
\toprule
&     & \multicolumn{3}{c}{\textbf{F1}}     &  & \multicolumn{2}{c}{\textbf{Macro-Recall}}  &  & \multicolumn{2}{c}{\textbf{Rouge-L}} &  & \multicolumn{2}{c}{\textbf{BLEU-1}}  &  & \multicolumn{2}{c}{\textbf{BERTScore}} \\ 
\cmidrule{3-5} \cmidrule{7-8} \cmidrule{10-11} \cmidrule{13-14} \cmidrule{16-17}
\multirow{-2}{*}{\textbf{Model}}   & \multirow{-2}{*}{\textbf{Settings}} & \textbf{Pre} & \textbf{DD} & \textbf{Fin} &  & \textbf{Pre}  & \textbf{Fin}              &  & \textbf{Pre}      & \textbf{Fin}     &  & \textbf{Pre}     & \textbf{Fin}    &  & \textbf{Pre}     & \textbf{Fin}   \\ \midrule
\multicolumn{17}{c}{\cellcolor[HTML]{EFEFEF}\textbf{\textit{Open-source Medical LLMs}}}    \\ \midrule
MMed-Llama-3-8B      & 1-Shot     & 22.03       & 8.87        & 18.19        &  & 40.73        & 32.68                    &  & 22.29             & 14.77           &  & 12.83            & 6.02      &  & 60.90            & 60.13       \\
PULSE-20bv5          & 1-Shot     & 26.40        & 9.55        & 13.18        &  & 27.57        & 24.37                     &  & 45.93            & 42.20            &  & 36.11           & 32.44     &  & 80.46            & 78.46       \\
Apollo2-7B        & 1-Shot    & 33.61        & 8.47        & 26.78         &  & 37.01         & 33.15    &  & 51.25             & 44.91             &  & 37.00             & 33.10           &  & 80.70      & 78.50  \\ \midrule
\multicolumn{17}{c}{\cellcolor[HTML]{EFEFEF}\textbf{\textit{Open-source General LLMs}}}          \\ \midrule
    & 1-Shot        & 32.72       & 5.54        & 19.99        &  & 37.88         & 22.11                     &  & 51.91             & 32.10          &  & 34.42            & 27.86       &  & 82.10      & 77.13    \\
\multirow{-2}{*}{Llama-3.1-8B}     & LoRA       & 20.59        & 7.86        & 16.32        &  & 17.04         & 14.06                     &  & 25.07             & 19.76            &  & 18.69            & 15.00        &  & 70.97      & 70.20    \\ \cmidrule{2-17} 
    & 1-Shot       & 34.71        & 9.97        & 31.76       &  & 38.53         & 40.48                    &  & 50.41             & 51.35          &  & 38.65            & 39.37        &  & {\ul 85.13}      & 85.53    \\
\multirow{-2}{*}{glm4-chat-9b}     & LoRA           & \textbf{38.78}        & 10.66      & {\ul 34.00}        &  & 42.36         & 42.89                     &  & {\ul 59.23}             & \textbf{58.51}            &  & 44.50             & {\ul 45.89}        &  & 84.96      &{\ul  85.73}    \\ \cmidrule{2-17} 
                                   & 1-Shot     & 24.77       & 2.05        & 15.83        &  & 29.67         & 15.49                     &  & 38.54             & 31.02          &  & 29.88            & 23.22         &  & 77.16      & 73.50   \\
\multirow{-2}{*}{Baichuan2-13B}    & LoRA     & 32.51        & \textbf{13.84}       & 28.25        &  & \textbf{51.15}         & \textbf{52.58}                     &  & \textbf{60.69}             & {\ul 58.07}            &  & 45.46            & 44.08       &  & 74.60      & 74.56       \\ \midrule
\multicolumn{17}{c}{\cellcolor[HTML]{EFEFEF}\textbf{\textit{Closed-source LLMs}}}           \\ \midrule
                                   & 1-Shot    & 30.93        & 8.08        & 20.99        &  & 39.29         & 31.46                     &  & 51.54            & 46.10            &  & {\ul 47.36}            & 40.82         &  & 82.16      & 81.46   \\
                                   & CoT       & 30.32        & 3.63        & 22.29        &  & 26.51         & 19.10                     &  & 38.65             & 34.18            &  & 29.47            & 25.91        &  & 72.36      & 70.89   \\
                                   & DAC       & 24.74       & 8.07        & 25.05        &  & 28.82         & 35.34                    &  & 40.40             & 49.64            &  & 29.88            & 42.37          &  & 81.17      & 82.35  \\
\multirow{-4}{*}{ChatGPT3.5-turbo} & Ours      & 34.60        & 8.23        & 23.29       &  & 39.38         & 39.02 &  & 51.54             & 47.19            &  & 47.04            & 45.05         &  & 82.27      & 83.31    \\ \cmidrule{2-17} 
                                   & 1-Shot    & 31.00           & 10.10       & 28.92        &  & 34.61         & 35.82                     &  & 53.28             & 53.61            &  & 44.91            & 45.45        &  & 84.39      & 84.56    \\
                                   & CoT      & 33.79        & 7.06       & 31.28        &  & 27.63         & 26.31                     &  & 48.95             & 48.67           &  & 38.66            & 37.55          &  & 78.63      & 79.03  \\
                                   & DAC      & 26.95        & 10.17      & 24.90        &  & 23.18         & 24.26                     &  & 52.33             & 51.96            &  & 40.42            & 39.86          &  & 80.83      & 81.13  \\
\multirow{-4}{*}{GPT4o-mini}       & Ours     & {\ul 34.78}       & {\ul 11.13}      &  \textbf{34.32}        &  & {\ul 42.67}         & {\ul 43.28}                     &  & 55.95             & 55.97            &  & \textbf{49.15}            & \textbf{47.94}         &  & \textbf{86.13}      & \textbf{86.06}   \\ \midrule
\multicolumn{17}{c}{\cellcolor[HTML]{EFEFEF}\textbf{\textit{Other Method}}}           \\ \midrule
Human     & -            & 94.11        & 84.31        & 95.31        &  & 96.08         & 97.16       &  & 97.64             & 97.44            &  & 97.27            & 97.41         &  & 98.03      & 98.19   \\ \bottomrule
\end{tabular}
\label{main_result}
\end{table*}

\subsection{Backward Inference and Reflection}


After obtaining the forward inference results, this part aims to conduct backward inference, reflection, and refine the patient’s diagnosis. In this section, we define some rules to achieve this goal.

Specifically, given the forward inference diagnostic results $a_o$, we first perform backward inference from the diagnosis to diagnostic criteria. In this step, we define backward inference rules to guide the LLM in generating outputs that comply with pre-defined content and format constraints. For content constraints, we have ``For each diagnosis, recall the representative medical history, symptoms, physical signs, and examination results'' (Rule 1). For format constraints, we consider ``The recalled content should follow the format:
Medical History: Recall the representative medical history for the disease; delete this item if not applicable'' (Rule 2).
Then, we design reflection rules to review the diagnoses. 
The specific reflection rule is: “If a diagnosis’s characteristics don’t align with the medical record, delete or revise it, and provide the rationale”.
Finally, we combine the aforementioned backward inference, reflection results, and their criteria to let the model optimize the diagnostic results, 
with specific prompts shown in Appendix~\ref{prompt}.

\section{Experiments}

In this section, we conduct a series of experiments to evaluate the performance of LLMs on the MSDiagnosis and the effectiveness of our proposed framework. Additionally, we provide an analysis of the multi-step diagnostic process and the proposed framework.

\subsection{Experimental Setup}

\subsubsection{Baseline}


In this paper, we mainly describe several types of baseline methods, including open-source medical LLMs, open-source general LLMs, closed-source LLMs, and other methods. 
The specific settings are shown in the Appendix~\ref{baseline}.


\subsubsection{Evaluation Details}

For different types of problems, we introduce various evaluation metrics. 
Specifically, for primary diagnosis, differential diagnosis, and final diagnosis, we introduce entity $F1$~\cite{entityf1} as the evaluation metric. 
For primary and final diagnosis criteria, we employ two types of evaluation metrics.
First, we adapt $Rouge-L$~\cite{rouge}, $BERTScore$~\cite{BERTScore}, and $BLEU$-1~\cite{blue} to measure the similarity between the generated text and the reference text.
Second, we introduce the $Macro-Recall$ metric, which is calculated based on the key points in the answers. 
More details are shown in Appendix~\ref{evaluation}.




\subsubsection{Implementation Details}


For all the open-source models mentioned above, we use their default hyperparameters. 
All experiments are conducted three times, and the average performance across three runs is computed.
Further implementation details are listed in Appendix~\ref{details}.

\begin{table*}[]
\small
\caption{The ablation results of the multi-step diagnostic (\%). \emph{w/o} represents deleting the corresponding process.}
\centering
\setlength{\tabcolsep}{1.5mm}
\begin{tabular}{lccccccccccccccc}
\toprule
& \multicolumn{3}{c}{\textbf{F1}}           &  & \multicolumn{2}{c}{\textbf{Macro-Recall}} &  & \multicolumn{2}{c}{\textbf{Rouge-L}} &  & \multicolumn{2}{c}{\textbf{BLEU-1}}  &  & \multicolumn{2}{c}{\textbf{BERTScore}} \\ 
\cmidrule{2-4} \cmidrule{6-7} \cmidrule{9-10} \cmidrule{12-13} \cmidrule{15-16} 
\multirow{-2}{*}{\textbf{Method}} & \textbf{Pre} & \textbf{DD} & \textbf{Fin} &  & \textbf{Pre}        & \textbf{Fin}        &  & \textbf{Pre}      & \textbf{Fin}     &  & \textbf{Pre}    & \textbf{Fin}  &  & \textbf{Pre}    & \textbf{Fin}  \\ \midrule
\multicolumn{16}{c}{\cellcolor[HTML]{EFEFEF}\textbf{\textit{Primary diagnosis matches final diagnosis}}}                                                                                                                                                             \\ \midrule
\emph{w/o} primary diagnosis  & -            & 8.05        & 44.46        &  & -                   & 42.66               &  & -                 & 59.53            &  & -               & 52.30      &  & -               & 86.40      \\
\emph{w/o} differential  diagnosis  & 40.8         & -           & 41.28        &  & 35.96               & 42.10               &  & 57.14             & 58.67            &  & 48.08           & 50.62     &  & 85.55               & 86.08       \\
multi-step     & 38.11        & 8.23        & 39.33        &  & 35.94               & 42.96               &  & 55.94             & 58.62            &  & 46.88           & 49.21        &  & 85.14               & 85.77    \\ \midrule
\multicolumn{16}{c}{\cellcolor[HTML]{EFEFEF}\textbf{\textit{Primary diagnosis differs from final diagnosis}}}                                                                                                                                                        \\ \midrule
\emph{w/o} primary diagnosis   & -            & 9.80        & 21.17        &  & -                   & 29.72               &  & -                 & 49.66            &  & -               & 42.35       &  & -               & 83.05     \\
\emph{w/o} differential  diagnosis    & 25.19        & -           & 18.78        &  & 32.70               & 30.08               &  & 50.28             & 49.04            &  & 43.69           & 41.91    &  & 83.62               & 83.03       \\
multi-step    & 23.99        & 11.49       & 18.12        &  & 32.50               & 30.91               &  & 50.74             & 48.76            &  & 42.90           & 41.88       &  & 83.46               & 83.09    \\ \bottomrule
\end{tabular}
\label{ablation_dataset}
\end{table*}

\begin{table*}[t]
\caption{The ablation results of our framework (\%). \emph{w/o}
indicates the removal of the corresponding module.}
\small
\centering
\setlength{\tabcolsep}{1.55mm}
\begin{tabular}{lccccccccccccccc}
\toprule
\multirow{2}{*}{\textbf{Method}} & \multicolumn{3}{c}{\textbf{F1}}             &  & \multicolumn{2}{c}{\textbf{Macro-Recall}} &  & \multicolumn{2}{c}{\textbf{Rouge-L}} &  & \multicolumn{2}{c}{\textbf{BLEU-1}}  &  & \multicolumn{2}{c}{\textbf{BERTScore}} \\ 
\cmidrule{2-4} \cmidrule{6-7} \cmidrule{9-10} \cmidrule{12-13} \cmidrule{15-16}
& \textbf{Pre}   & \textbf{DD}    & \textbf{Fin}   &  & \textbf{Pre}        & \textbf{Fin}        &  & \textbf{Pre}      & \textbf{Fin}     &  & \textbf{Pre}    & \textbf{Fin}   &  & \textbf{Pre}    & \textbf{Fin}  \\ \midrule
\emph{w/o} Backward inference    & 34.59          & 10.96          & 32.04          &  & 41.32        & 41.01        &  & 55.48             & 55.48            &  & 48.93           & 47.32      &  & 85.83           & 84.76    \\
\emph{w/o} Reflection            & 34.68          & 10.04          & 32.19          &  & 41.08        & 41.60        &  & 55.53             & 55.89            &  & 49.16           & 47.87      &  & 85.93           & 85.93     \\
\emph{w/o} Refinement            & 34.63          & 10.81          & 32.92          &  & 41.95        & 42.15        &  & 55.63             & 55.39            &  & 49.06           & 47.72      &  & 85.16           & 84.83      \\  \midrule
Ours                             & \textbf{34.78} & \textbf{11.13} & \textbf{34.32} &  & \textbf{42.67}      & \textbf{43.28}      &  & \textbf{55.95}    & \textbf{55.97}   &  & \textbf{49.15}  & \textbf{47.94}  &  & \textbf{86.13}      & \textbf{86.06}  \\ \bottomrule
\end{tabular}
\label{ablation}
\end{table*}

\subsection{Main Results}

To verify the effectiveness of our proposed method, we compare it with all baselines on the MSDiagnosis test set. The results are reported in Table~\ref{main_result}.

From the table, we conclude that:
1) All LLMs perform poorly on MSDiagnosis, with a significant gap from the human final diagnosis $F1$ of 95.31\%.
Specifically, for the final diagnosis question, the best-performing model, glm4 with LoRA, achieves a final $F1$ of 38.78\%. 
2) Our framework outperforms all baselines in the final diagnosis $F1$ and $Blue$-1 metrics, demonstrating the effectiveness of the proposed framework.
In particular, our framework improves the final $F1$ by 0.32\% over glm4 with LoRA.
3) Among open-source general LLMs, the glm4 and Baichuan2 models outperform those without instruction fine-tuning in diagnostic reasoning after fine-tuning. 
Specifically, after fine-tuning, glm4 improves by 4.07\% and 2.24\% in primary $F1$ and final $F1$ metrics, respectively. 
However, the LlaMA’s performance declines after fine-tuning, with primary $F1$ and final $F1$ metrics decreasing by 12.13\% and 3.67\%, respectively. This decline likely results from the relatively small proportion of Chinese in LlaMA’s training corpus. Using Chinese data for instruction fine-tuning could further degrade its performance.
4) The model performs worse in differential diagnosis than in primary and final diagnoses. This is due to the increased complexity of the differential diagnosis task, which requires distinguishing between similar diseases and demands more specialized medical knowledge.

\subsection{Detailed Analysis}

\definecolor{c1}{HTML}{793DA5}
\definecolor{c2}{HTML}{FBC000}
\definecolor{c3}{HTML}{2AB051}
\definecolor{c4}{HTML}{2070C0}

\begin{figure*}[ht]
\centering
\includegraphics[width=2\columnwidth]{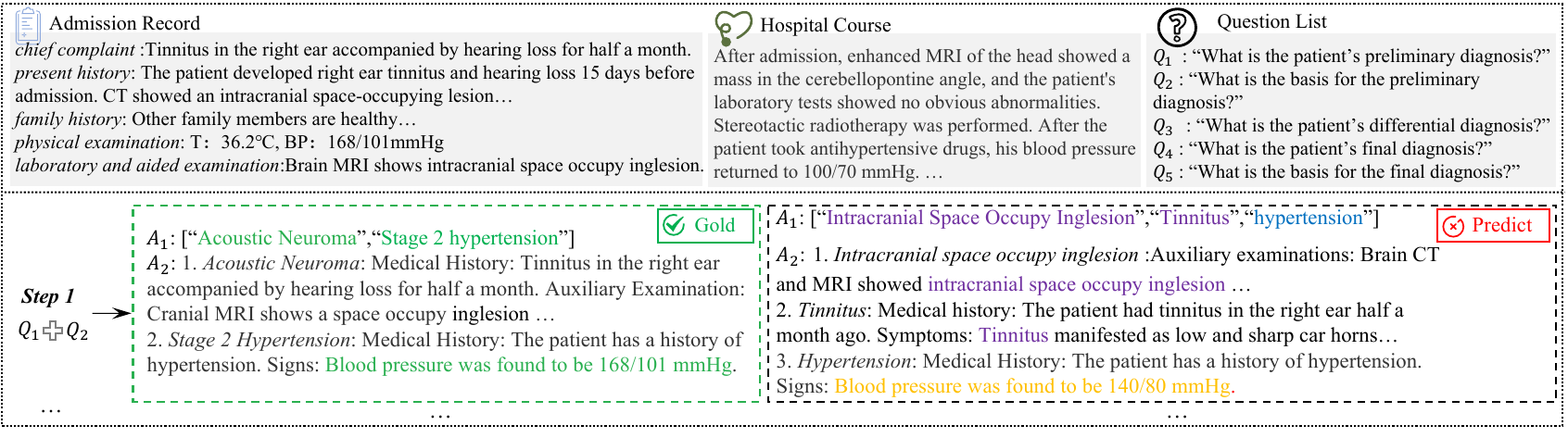} 
\caption{Case study. The \textcolor{c3}{green} (\textcolor{red}{red}) highlight indicates correct (incorrect) results. \textcolor{c1}{Purple} marks lack of domain knowledge errors, \textcolor{c4}{blue} marks symptom-disease confusion errors, and \textcolor{c2}{yellow} marks diagnostic criteria inconsistent with facts.}
\label{new_case}
\end{figure*}

\subsubsection{Ablation Study}


In this section, we design two ablation experiments to analyze the multi-step diagnostic process of MSDiagnosis and our framework respectively. 
Specifically, for the ablation of the multi-step diagnostic process, we sequentially remove the primary diagnosis, differential diagnosis, and both to evaluate the effectiveness of the final diagnosis. 
Notably, the LLM is used for direct reasoning in this experiment.
The results are shown in Table~\ref{ablation_dataset}. 
For the ablation of our framework, we sequentially remove backward inference, reflection, and refinement, assessing the effectiveness of the remaining components. The results are presented in Table~\ref{ablation}.

From the Table~\ref{ablation_dataset}, we conclude that: 
1) The primary diagnosis significantly influences the final diagnosis's performance, likely due to errors in the primary diagnosis propagating through the multi-step process, reducing the final diagnosis’s effectiveness.
2) The multi-step diagnostic process helps improve the interpretability of the final diagnosis. Specifically, by introducing a multi-step diagnosis, the final diagnosis is based on more criteria, thereby enhancing its interpretability.
From the Table~\ref{ablation}, it can be observed that when backward reasoning is removed, the $F1$ and $Macro-Recall$ scores for the final diagnosis decrease significantly, specifically by 2.28\% and 2.27\%, respectively. 
These findings suggest that backward reasoning effectively enhances both the accuracy and interpretability of the final diagnosis.


\subsubsection{Case Study}


To further investigate the limitations of our method, we analyze 100 error samples. After manual classification, the errors are categorized into three types: 
(a) lack of domain knowledge (42\%), (b) Symptoms or examination results are confused with the disease. (30\%), (c) diagnostic criteria inconsistent with facts (18\%), along with other errors (10\%). 
To illustrate these error types more intuitively, we provide examples as shown in Fig.~\ref{new_case}.

\section{Related Work}

\subsection{Clinical Diagnosis on EMRs}


Clinic diagnosis based on EMRs is crucial for improving healthcare quality and patient outcomes. However, existing datasets for EMR-based diagnosis primarily focus on single-step diagnosis, with diagnostic types generally divided into two categories: primary diagnosis and differential diagnosis.
Primary diagnosis involves determining the likely disease based on the patient’s history and symptoms, or in combination with examination results. Most current research focuses on extracting abnormal features from medical records or integrating external medical knowledge to make a diagnosis~\cite{ram, medikal, realm}.
Differential diagnosis is a list of potential diseases that could cause the patient’s symptoms~\cite{ddx}. It enables a more comprehensive evaluation of clinical EMRs, allowing for the identification of less obvious but critical conditions. Current studies often employ deep learning methods to extract features from medical records for differential diagnosis~\cite{dualddx, wuddx}.
In summary, previous research on EMR-based diagnosis has primarily focused on single-step diagnostic processes. However, this approach does not align with actual clinical diagnostic workflows. 

\subsection{Prompting Strategies of LLM}

With the development of LLMs, many researchers have applied these models to medical tasks. These methods can be categorized into three types: IO prompting, CoT prompting, and the DAC paradigm. 
IO prompting~\cite{IO} is a standard prompting strategy where input is combined with instructions and/or a few ICLs to generate a response. 
CoT prompting~\cite{cot} aims to emulate the step-by-step thought process humans use to tackle complex tasks, such as combinatorial reasoning and mathematical calculations. There are also various CoT variants, such as CoT with self-consistency (CoT-SC) prompting~\cite{cotsc}, designed to address the limitations of CoT in exploration. 
The DAC paradigm~\cite{DAC} mainly refers to simply breaking down the input sequence into multiple sub-inputs to enhance LLM performance on certain specific tasks. 
While these methods show promise in reasoning tasks, they are challenging to apply directly to medical diagnostic reasoning.

\section{Conclusion}


This paper introduces MSDiagnosis, a multi-step clinical diagnostic benchmark that is collected and annotated from medical websites.
MSDiagnosis addresses the limitations of existing datasets by constructing multi-step diagnostic tasks that better align with actual clinical diagnostic scenarios. 
For this benchmark, we propose a simple and effective framework. 
We implement both closed-source and open-source LLMs and conduct extensive experiments. 
The results show that tasks in our benchmark effectively measure the multi-step clinical diagnostic abilities, and the framework proposed in this paper shows effectiveness on this benchmark.

\section*{Limitations}
The limitations of our work are as follows.
Firstly, due to limited data sources, our medical records dataset exhibits an uneven distribution across different departments. This issue can be addressed through machine learning methods~\cite{learning} and data sampling strategies~\cite{smote}.
Secondly, the proposed framework requires multiple uses of the LLM, resulting in a longer inference time for multi-step diagnostic reasoning compared to direct reasoning.

\section*{Ethical Statement}



The raw data used in this study comes from an open-source medical platform, which includes a large volume of EMRs.
The platform permits data for research and education, as confirmed by prior studies~\cite{MEDIQA, automir, dermavqa}. 
Throughout the data collection process, we strictly ensured the avoidance of copyright and privacy issues. Furthermore, we conducted a thorough review of the dataset to ensure it contains no harmful content, such as gender bias, racial discrimination, or other inappropriate materials.


\bibliography{ref}

\clearpage
\appendix


\definecolor{c7}{HTML}{FBE2E5}
\definecolor{c8}{HTML}{E6FFE6}

\begin{table*}[ht]
\small
\caption{The impact of different departments on our framework. \colorbox{c8}{Green} highlights indicate that our framework performs well in processing medical records for the respective department, while \colorbox{c7}{red} highlights indicate poorer performance in that department.}
\centering
\setlength{\tabcolsep}{1.2mm}
\begin{tabular}{lccccccccccccccc}
\toprule
\multirow{2}{*}{\textbf{Department}} & \multicolumn{3}{c}{\textbf{F1}}           &  & \multicolumn{2}{c}{\textbf{Macro-Recall}} &  & \multicolumn{2}{c}{\textbf{Rouge-L}} &  & \multicolumn{2}{c}{\textbf{BLEU-1}} &  & \multicolumn{2}{c}{\textbf{BERTScore}}\\ 
\cmidrule{2-4} \cmidrule{6-7} \cmidrule{9-10} \cmidrule{12-13}  \cmidrule{15-16} 
& \textbf{Pre} & \textbf{DD} & \textbf{Fin} &  & \textbf{Pre}        & \textbf{Fin}        &  & \textbf{Pre}      & \textbf{Fin}     &  & \textbf{Pre}    & \textbf{Fin} &  & \textbf{Pre}    & \textbf{Fin}    \\ \midrule
Pediatrics  & 37.58        & 10.16       & 35.84        &  & 49.54               & 47.35               &  & 52.03             & 52.27            &  & 45.75           & 46.80     &  & 84.56    & 84.19  \\
Internal Medicine   & 27.09        & 14.30       & 22.10        &  & 39.08         & 36.63               &  & 51.10          & 52.14         &  & 47.35      & 45.95    &  & 84.74       & 84.38    \\
Emergency   & 32.14        & 29.64       & 32.14        &  & 44.91               & 42.88               &  & 61.09             & 57.71            &  & 53.15         & 58.51   &  & 87.66    & 88.06  \\
Nursing Department  & 54.60        & 11.11       & 45.71        &  & 29.16         & 36.66        &  & 62.24        & 63.70       &  & 51.34       & 43.09    &  & 85.87    & 85.76 \\
Surgery     & 40.68        & 10.40       & 39.81        &  & 43.21               & 47.65         &  & 60.93             & 61.17     &  & 52.25   & 50.93  &  & 87.40    & 87.00  \\
Obstetrics-Gynecology   & 37.14        & 22.44       & 23.08        &  & 32.00        & 29.41      &  & 49.97             & 47.51            &  & 49.65     & 40.76   &  & 85.34    & 84.02  \\
Otolaryngology (ENT)    & 44.82        & 5.64        & 41.83        &  & 44.42        & 50.28       &  & 63.23      & 63.60       &  & 49.37    & 47.78     &  & 87.53    & 87.53    \\
Psychiatry   & 33.33        & 0.00        & 22.22        &  & 19.97               & 31.97               &  & 69.46             & 50.77            &  & 43.16           & 46.91      &  & 84.30    & 83.45     \\
{\cellcolor[HTML]{FBE2E5}}Oncology     & {\cellcolor[HTML]{FBE2E5}}7.63         & {\cellcolor[HTML]{FBE2E5}}2.57        & {\cellcolor[HTML]{FBE2E5}}5.89         & {\cellcolor[HTML]{FBE2E5}} & {\cellcolor[HTML]{FBE2E5}}24.67               & {\cellcolor[HTML]{FBE2E5}}25.85               & {\cellcolor[HTML]{FBE2E5}} & {\cellcolor[HTML]{FBE2E5}}39.83             & {\cellcolor[HTML]{FBE2E5}}40.80             & {\cellcolor[HTML]{FBE2E5}} & {\cellcolor[HTML]{FBE2E5}}40.90          & {\cellcolor[HTML]{FBE2E5}}42.48     & {\cellcolor[HTML]{FBE2E5}} & {\cellcolor[HTML]{FBE2E5}}82.34    & {\cellcolor[HTML]{FBE2E5}}83.06      \\
Ophtalmology     & 20.83        & 7.14        & 20.83        &  & 44.01               & 36.46               &  & 55.66             & 53.64            &  & 47.56       & 48.08    &  & 86.25    & 86.92       \\
Dermatology      & 83.33        & 0.00        & 75.00        &  & 38.54               & 48.96               &  & 33.93             & 67.86            &  & 52.08       & 45.76     &  & 86.65    & 88.36       \\
{\cellcolor[HTML]{E6FFE6}Stomatology}      & {\cellcolor[HTML]{E6FFE6}75.00}        & {\cellcolor[HTML]{E6FFE6}8.33}        & {\cellcolor[HTML]{E6FFE6}83.33}        &  {\cellcolor[HTML]{E6FFE6}} & {\cellcolor[HTML]{E6FFE6}48.87}               & {\cellcolor[HTML]{E6FFE6}62.07}               & {\cellcolor[HTML]{E6FFE6} } & {\cellcolor[HTML]{E6FFE6}74.22}             & {\cellcolor[HTML]{E6FFE6}85.90}           &  {\cellcolor[HTML]{E6FFE6}} & {\cellcolor[HTML]{E6FFE6}56.57}      & {\cellcolor[HTML]{E6FFE6}64.02}      &  {\cellcolor[HTML]{E6FFE6}} & {\cellcolor[HTML]{E6FFE6}88.50}   & {\cellcolor[HTML]{E6FFE6}93.03}     \\
Traditional Chinese Medicine   & 0.00         & 40.0    & 0.00   &  & 61.88      & 43.75     &  & 83.33      & 53.33    &  & 58.95    & 42.38       &  & 88.63    & 83.80     \\
Others  & 38.76        & 10.97       & 23.82        &  & 37.87               & 36.34               &  & 62.97             & 61.62            &  & 50.66           & 47.16    &  & 87.07    & 85.68        \\ \bottomrule
\end{tabular}
\label{depart_impart}
\end{table*}

\section{Details of the Baseline}
\label{baseline}
In this paper, we mainly describe several types of baseline methods, including open-source medical LLMs, open-source general LLMs, closed-source LLMs, and other method. 

In the open-source medical LLMs, we employ MMedLM~\cite{mmed-llama}, PULSE~\cite{pulse}, and Apollo2-7B~\cite{Apollo2-7B} for comparison. 
Based on these models, we manually construct an example to serve as ICL. The complete example is shown in Section~\ref{complete_example}.
We previously tested several medical models (HuatuoGPT2-7B~\cite{huatuogpt}, DoctorGLM~\cite{doctorglm}, and BianQue-2~\cite{bianque}), but they were excluded due to poor performance, as the length of our dataset approached their context window limits.
According to our statistics, in the diagnostic tasks of this study, the average number of context tokens for the primary diagnosis is 3245.14, for the differential diagnosis is 3336.75, and for the final diagnosis is 4807.20.

In the open-source general LLMs, we use Llama-3.1~\cite{llama}, glm4~\cite{glm}, and Baichuan2~\cite{baichuan} for comparison. Based on these models, we design two settings: 1-shot reasoning and instruction tuning. 
In the first setting, we utilize the previously constructed example as ICL.
In the second setting, we use parameter-efficient fine-tuning with LoRA~\cite{lora}. The instruction data is generated by transforming the input and output from the training data.

In the closed-source LLMs, we use ChatGPT3.5-turbo\footnote{\url{https://platform.openai.com/docs/models/gpt-3-5-turbo}} and GPT4o-mini~\cite{gpt4} for comparison. We are unable to include this for all MSDiagnosis benchmarks due to the extraordinarily high cost of o1-preview inference.
Based on these models, we employ two settings: 1-shot and prompting methods.
In the first approach, we utilize the same example as previously used.
In the second setting, we consider comparing method COT~\cite{cot} and method DAC~\cite{DAC}.
COT~\cite{cot} uses ``Let’s think step by step'' to enhance the model's reasoning ability for task-solving. DAC~\cite{DAC} adopts a simple divide-and-conquer prompting strategy, where the input sequence is simply divided into multiple sub-inputs, which can enhance the reasoning performance of the LLM.

In other method, we mainly consider manual answering methods. Specifically, we randomly select 100 samples from MSDiagnosis and invite one college student (different from the annotation team in Section~\ref{answer_annotation}) to answer the questions.
Our evaluation method for human results is consistent with other baselines, utilizing the automated assessment approach described in Section~\ref{evaluation}.

\section{Details of the Evaluation Metrics}
\label{evaluation}
For the disease entities in the diagnosis results $D$ and the reference results $R$ in the medical records, we employ a more rigorous evaluation method. This approach aligns better with actual clinical needs and facilitates targeted treatment for patients.
Specifically, we compute the edit distance to associate these entities with ICD-10 terms, thereby mapping $D$ and $R$ to two standardized disease sets, $S_d$ and $S_r$, respectively. We then compute the entity $F1$ score based on $S_d$ and $S_r$.

For primary and final diagnosis criteria, we employ two types of evaluation metrics.
First, we adapt $Rouge-L$~\cite{rouge}, $BERTScore$~\cite{BERTScore}, and $BLEU$-1~\cite{blue} to measure the longest matching sequence between the generated text and the reference text, capturing the similarity in the overall structure of the two sequences.
Second, we introduce the $Macro-Recall$ metric, which is calculated based on the key points in the answers. 
For our predefined $N$ key point categories, the calculation method for key point $Macro-Recall$ is as follows:
\begin{equation}
Marco-Recall=\frac1N \sum_{i=1}^N Recall_i,
\end{equation}
where $Recall_i$ represents the recall rate of the $i$ category.

\section{Details of Implementation}
\label{details}
To enhance the stability and reliability of the experimental results and reduce the impact of random factors, we conduct each experiment three times and then calculate the average of three results.
For the backbone model, we utilize the OpenAI API, specifying the model as ``GPT4o-mini''. We set the $top\_p$ parameter to 0.01, and all other hyperparameters of the OpenAI API are maintained at default values. When selecting similar examples, we set $K$ to 1.
During the LoRA fine-tuning phase, we employ the Adam optimizer with weight decay correction for fine-tuning the model. The initial learning rate is set to 1e-4, and the batch size is set to 1, with Cross-Entropy Loss serving as the loss function.
The input during training is: "instruction: {task description}. Medical Record: {patient case}. question: {diagnosis question}". The output is: "the answer to the question". We calculate the loss only for the answer, excluding the instruction and input.
For the open-source models, our experiments are conducted on four Nvidia A100 GPUs, each with 40GB of memory, and we use PyTorch\footnote{\url{https://pytorch.org/}} in Python\footnote{\url{https://www.python.org/}}.

\section{Question Definition}
\label{question_denfinition}

\begin{table}[t]
\small
\caption{The definition of question}
\setlength{\tabcolsep}{0.9mm}
\begin{tabular}{ll}
\toprule
\textbf{Question} & \textbf{Definition}                                                                 \\ \midrule
$Q_1$       & Inquire about the patient’s primary diagnosis. \\ \midrule
$Q_2$       & \makecell[l]{Inquire about the patient’s primary \\ diagnostic criteria.} \\ \midrule
$Q_3$       & Inquire about the patient’s differential diagnosis. \\ \midrule
$Q_4$       & Inquire about the patient’s final diagnosis.  \\ \midrule
$Q_5$       & \makecell[l]{Inquire about the patient’s final\\ diagnostic criteria.} \\ 
\bottomrule
\end{tabular}
\label{question}
\end{table}

In this section, we introduce in detail the definition of the diagnostic questions corresponding to each EMR. In MSDiagnosis, the questions corresponding to each case are initially constructed manually and then expanded using GPT-4. The definitions of each question are shown in Table~\ref{question}.

\begin{table*}[t]
\small
\caption{The impact of the number of diagnosed diseases (final diagnosis) on the framework.}
\centering
\setlength{\tabcolsep}{1.75mm}
\begin{tabular}{lccccccccccccccc}
\toprule
\multirow{2}{*}{\textbf{Number of Disease}} & \multicolumn{3}{c}{\textbf{F1}}           &  & \multicolumn{2}{c}{\textbf{Macro-Recall}} &  & \multicolumn{2}{c}{\textbf{Rouge-L}} &  & \multicolumn{2}{c}{\textbf{BLEU-1}} &  & \multicolumn{2}{c}{\textbf{BERTScore}}\\ 
\cmidrule{2-4} \cmidrule{6-7} \cmidrule{9-10} \cmidrule{12-13} \cmidrule{15-16} 
& \textbf{Pre} & \textbf{DD} & \textbf{Fin} &  & \textbf{Pre}        & \textbf{Fin}        &  & \textbf{Pre}      & \textbf{Fin}     &  & \textbf{Pre}    & \textbf{Fin} &  & \textbf{Pre}    & \textbf{Fin} \\ \midrule
1-5  & 35.93        & 11.63       & 32.50        &  & 42.52               & 45.73        &  & 57.33      & 58.4      &  & 50.17     & 48.62     &  & 86.32     & 86.23  \\
5-10 & 30.03        & 9.77        & 26.46        &  & 31.73               & 22.47        &  & 49.76      & 45.92     &  & 45.39     & 45.16     &  & 84.58     & 84.75  \\
\textgreater{}10    & 34.27       & 9.35         & 29.18      &  & 17.94      & 22.57    &  & 48.82      & 50.19     &  & 45.95     & 42.86     &  & 84.66     & 86.17  \\ \bottomrule
\end{tabular}
\label{diagnosis_impact}
\end{table*}

\section{The Impact of Different Departments on the Framework}


In this section, to evaluate the impact of different departments on our framework, we conduct an analysis of the department-specific results. The detailed experimental outcomes are presented in Table~\ref{depart_impart}. From this analysis, we can draw the following conclusions:
Different departments have different influences on the framework. 
The Oncology have the greatest impact on our framework.
Specifically, the final $F1$ score in the Oncology department is only 5.89\%. 
The best performance is achieved in the Stomatology department. Specifically, the final $F1$ score in the Oncology department is only 83.33\%.

\section{The Impact of the Number of Diagnosed Diseases on the Framework}


In this section, we primarily analyze the impact of the number of diseases on the framework. 
In this experiment, we categorize the number of diseases into three levels: 1 to 5 diseases, 5 to 10 diseases, and more than 10 diseases. 
The specific experimental results are shown in Table~\ref{diagnosis_impact}. 
From the results, we observe that as the number of diseases increases, the performance of the framework gradually declines. 
Specifically, when comparing patients with 5 to 10 diseases to those with 1 to 5 diseases, the final $F1$ score decreases by 6.04\%. 
However, when comparing patients with more than 10 diseases to those with 5 to 10 diseases, the final $F1$ score improves by 2.72\%. 
This improvement might be due to the fact that there are fewer patients with more than 10 diseases. 
Our analysis shows that there are only 5 such patients.

\begin{table*}[h]
\small
\caption{Experimental results with different numbers of ICL examples.}
\centering
\setlength{\tabcolsep}{2.3mm}
\begin{tabular}{lccccccccccccccc}
\toprule
\multirow{2}{*}{\textbf{Model}} & \multicolumn{3}{c}{\textbf{F1}}           &  & \multicolumn{2}{c}{\textbf{Macro-Recall}} &  & \multicolumn{2}{c}{\textbf{Rouge-L}} &  & \multicolumn{2}{c}{\textbf{BLEU-1}} &  & \multicolumn{2}{c}{\textbf{BERTScore}} \\ 
\cmidrule{2-4} \cmidrule{6-7} \cmidrule{9-10} \cmidrule{12-13} \cmidrule{15-16} 
& \textbf{Pre} & \textbf{DD} & \textbf{Fin} &  & \textbf{Pre}        & \textbf{Fin}    &  & \textbf{Pre}   & \textbf{Fin}  &  & \textbf{Pre}   & \textbf{Fin}  &  & \textbf{Pre}   & \textbf{Fin}  \\ \midrule  
ICL=0      & 34.70        & 10.09       & 31.91        &  & 34.58         & 31.91       &  & 54.25     & 53.31       &  & 47.04     & 44.66    &  & 84.30     & 84.45       \\
ICL=1      & 34.78        & 10.97       & 31.69        &  & 40.67         & 42.28       &  & 55.95     & 55.97       &  & 49.15     & 47.94    &  & 85.19   &  85.72      \\
ICL=2      & 34.76        & 10.72       & 31.53        &  & 41.60         & 43.88       &  & 56.92     & 55.28       &  & 50.42     & 48.72    &  & 86.45     & 86.04       \\
ICL=3      & 35.37        & 10.73       & 31.69        &  & 42.45         & 43.08       &  & 54.16     & 54.94       &  & 50.49     & 49.31    &  &  87.19  & 87.72 \\ \bottomrule
\end{tabular}
\label{icl}
\end{table*}

\section{The Impact of the Number of ICL on the Framework}

In this section, to analyze the impact of the number of ICL examples on the method’s performance, we introduced varying numbers of ICL examples into the framework for comparison. In this experiment, when the number of ICL examples reaches 4, the context length exceeds the model’s token limit, so we compared experiments with fewer than 4 ICL examples. The specific experimental results are shown in Table~\ref{icl}. As observed from the table, the performance of the framework gradually improves as the number of ICL examples increases.




\begin{figure}[ht]
\centering
\includegraphics[width=\columnwidth]{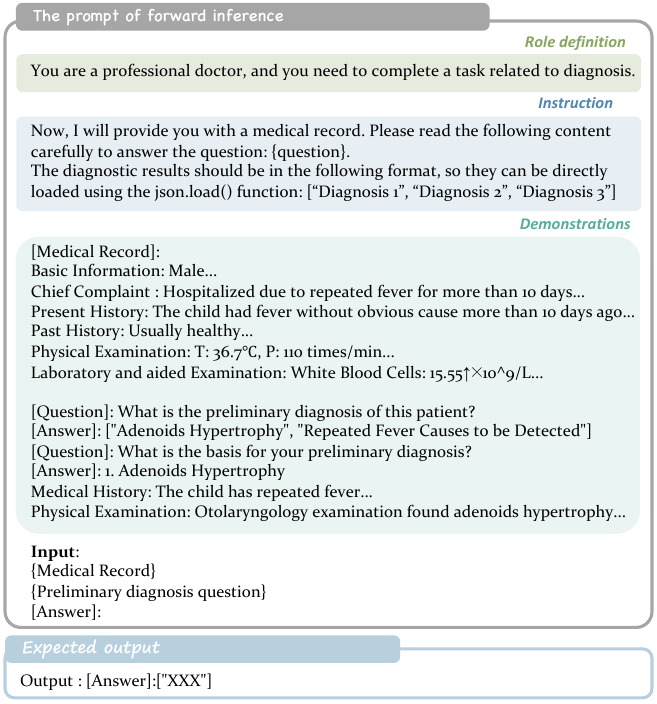} 
\caption{The prompt of forward inference.}
\label{zhengxiang}
\end{figure}

\begin{figure}[ht]
\centering
\includegraphics[width=\columnwidth]{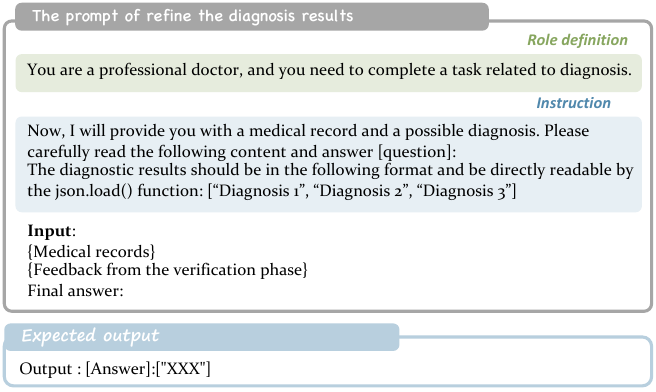} 
\caption{The prompt of refine the diagnosis results.}
\label{final_prompt}
\end{figure}

\begin{figure}[ht]
\centering
\includegraphics[width=\columnwidth]{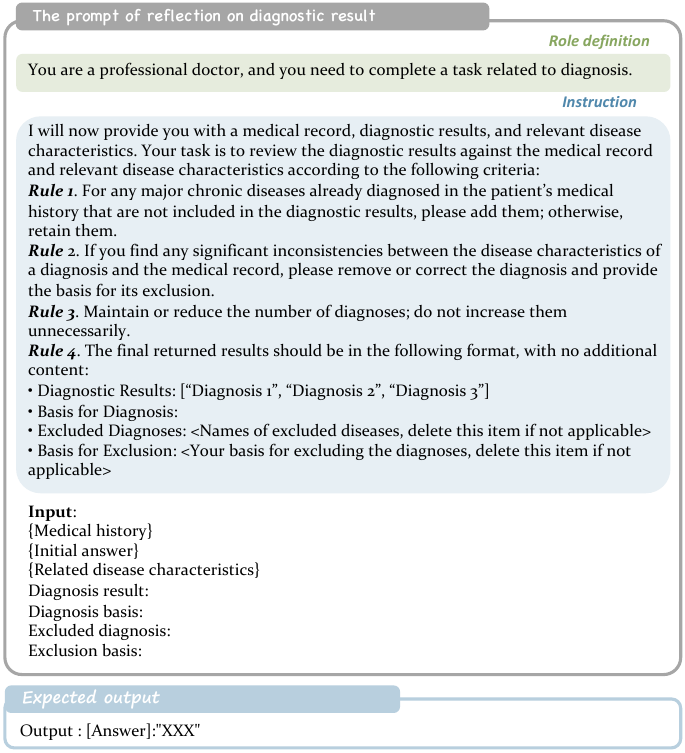} 
\caption{The prompt of reflection on diagnostic result.}
\label{yanzheng}
\end{figure}

\section{The Impact of Different Parameters of the Model on the Framework}


To evaluate the impact of models with varying parameter scales on the framework's performance, we evaluate Qwen2.5 models with 7B, 14B, 32B, and 72B parameters, and also compare with the 1-shot method for analysis convenience.
The detailed experimental results are presented in Table~\ref{different_model}. From the table, the following conclusions emerge: as the number of parameters increases, the framework’s performance improves progressively. Specifically, the 32B model outperforms the 7B model in the F1 and Macro-Recall metrics for final diagnosis by 2.18\% and 6.96\%, respectively. In comparison to the 14B model, the 32B model achieves improvements of 1.26\% and 1.12\% in these metrics.

\begin{table*}[t]
\setlength{\tabcolsep}{1.2mm}
\small
\caption{Comparison of models with different parameter sizes on our framework.}
\begin{tabular}{lllllllllllllllll}
\toprule
\multicolumn{1}{c}{\multirow{2}{*}{\textbf{Model}}} & \multicolumn{1}{c}{\multirow{2}{*}{\textbf{Method}}} & \multicolumn{3}{c}{\textbf{F1}}                                                                       & \multicolumn{1}{c}{} & \multicolumn{2}{c}{\textbf{Macro-Recall}}                           & \multicolumn{1}{c}{} & \multicolumn{2}{c}{\textbf{Rouge-L}}                                & \multicolumn{1}{c}{} & \multicolumn{2}{c}{\textbf{BLEU-1}}                                   &  & \multicolumn{2}{c}{\textbf{BERTScore}}                              \\ \cmidrule{3-5} \cmidrule{7-8} \cmidrule{10-11} \cmidrule{13-14} \cmidrule{16-17} 
\multicolumn{1}{c}{}                                & \multicolumn{1}{c}{}                                 & \multicolumn{1}{c}{\textbf{Pre}} & \multicolumn{1}{c}{\textbf{DD}} & \multicolumn{1}{c}{\textbf{Fin}} & \multicolumn{1}{c}{} & \multicolumn{1}{c}{\textbf{Pre}} & \multicolumn{1}{c}{\textbf{Fin}} & \multicolumn{1}{c}{} & \multicolumn{1}{c}{\textbf{Pre}} & \multicolumn{1}{c}{\textbf{Fin}} & \multicolumn{1}{c}{} & \multicolumn{1}{c}{\textbf{Pre}} & \multicolumn{1}{c}{\textbf{Fin}} &  & \multicolumn{1}{c}{\textbf{Pre}} & \multicolumn{1}{c}{\textbf{Fin}} \\ \midrule
\multirow{2}{*}{Qwen2.5-7B-Instruct}                & 1-Shot                                               & 34.65                            & 10.56                           & 29.97                            & \multicolumn{1}{c}{} & 34.90                            & 36.07                            & \multicolumn{1}{c}{} & 53.19                            & 54.41                            & \multicolumn{1}{c}{} & 41.73                            & 42.26                            &  & 86.67                            & 87.21                            \\
                                                    & Our                                                  & 36.97                            & 9.01                            & 33.35                            &                      & 39.39                            & 40.23                            &                      & 58.60                            & 59.81                            &                      & 50.32                            & 49.25                            &  & 87.61                            & 87.57                            \\ \cmidrule{2-17}
\multirow{2}{*}{Qwen2.5-14B-Instruct}               & 1-Shot                                               & 36.88                            & 11.72                           & 33.82                            &                      & 43.12                            & 43.18                            &                      & 60.27                            & 60.61                            &                      & 53.06                            & 50.48                            &  & 88.68                            & 89.84                            \\
                                                    & Our                                                  & 36.95                            & 12.41                           & 34.27                            &                      & 44.31                            & 46.07                            &                      & 63.07                            & 61.04                            &                      & 53.17                            & 52.01                            &  & 89.03                            & 89.56                            \\ \cmidrule{2-17}
\multirow{2}{*}{Qwen2.5-32B-Instruct}   & 1-Shot     & 37.43         & 11.92         & 33.40        &     & 43.46     & 45.82      &      & 52.99    & 51.16  &  & 38.07     & 36.45      &  & 85.72      & 85.61        \\
                                        & Our        & 38.27         & 11.79         & 35.53        &     & 45.24     & 47.19      &      & 62.83    & 63.38  &  & 55.43     & 52.64   &  & 88.37          & 88.48  \\ 
\cmidrule{2-17}
\multirow{2}{*}{Qwen2.5-72B-Instruct}   & 1-Shot     & 36.87         & 10.56         & 34.11        &     & 47.41     & 50.01      &      & 59.34    & 60.11  &  & 52.51     & 52.24      &  & 87.27      & 87.83        \\
                                        & Our        & 38.76         & 10.56         & 34.97        &     & 49.24     & 51.09      &      & 64.71    & 65.10  &  & 54.14     & 53.64      &  & 87.52      & 88.83  \\ 
                                        \bottomrule
\end{tabular}
\label{different_model}
\end{table*}

\section{The Prompt used in our Framework}
\label{prompt}
\begin{figure}[ht]
\centering
\includegraphics[width=\columnwidth]{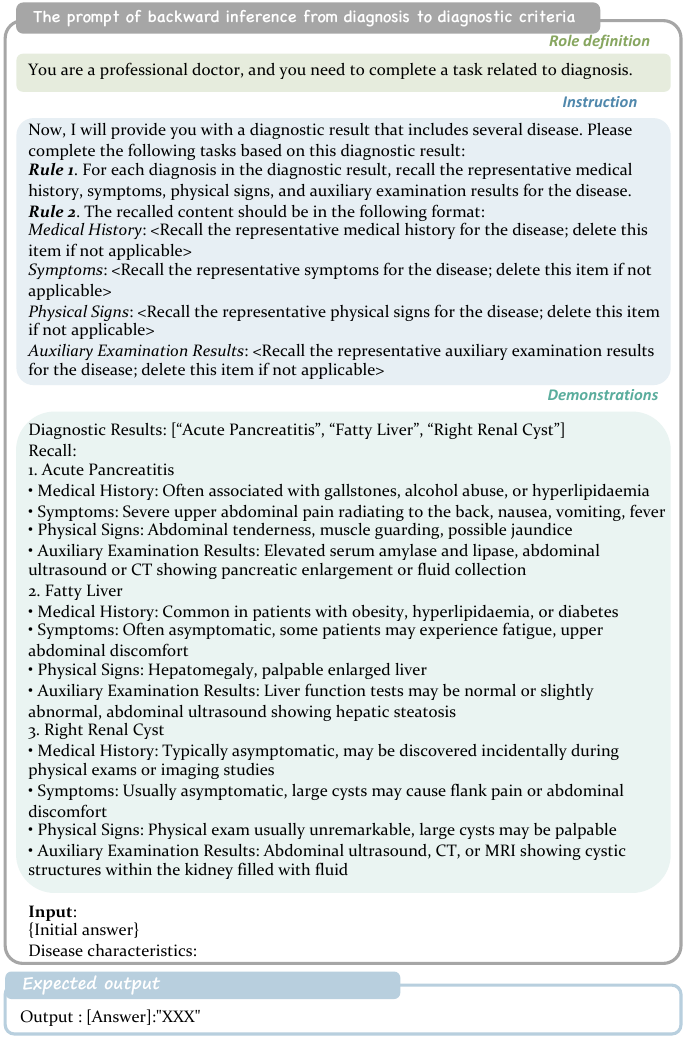} 
\caption{The prompt of backward inference from diagnosis to diagnostic criteria.}
\label{fanxiang}
\end{figure}

In this section, we primarily introduce the prompts corresponding to the four instances where the LLM is used within the framework.
It is noteworthy that the prompts in our framework follow a general pattern, including role definition, formatting instructions, and examples, rather than being meticulously designed.
The prompt for forward reasoning is shown in Fig.~\ref{zhengxiang}. The prompt for backward reasoning is shown in Fig.~\ref{fanxiang}. The prompt for reflecting on the diagnostic results is shown in Fig.~\ref{yanzheng}. The prompt for optimizing the diagnostic results is shown in Fig.~\ref{final_prompt}.

\section{Complete Multi-step Diagnostic Example}
\label{complete_example}

\begin{figure*}[ht]
\centering
\includegraphics[width=2\columnwidth]{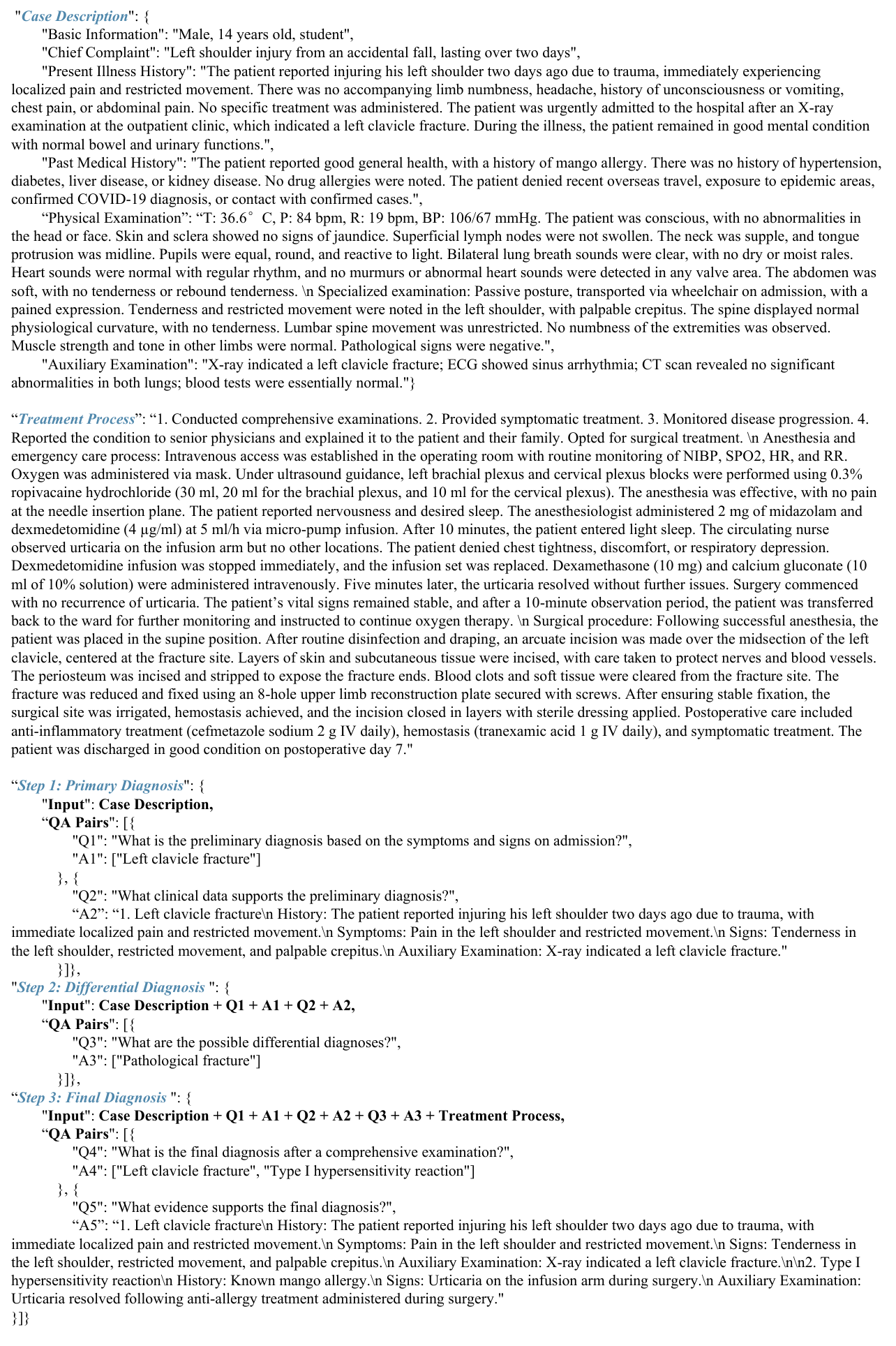} 
\caption{A complete multi-step diagnostic example.}
\label{fullexample}
\end{figure*}

We present a comprehensive multi-step diagnostic example, as illustrated in Fig.~\ref{fullexample}. In the first stage, the input is the medical record introduction, and Q1 and Q2 are answered sequentially. The responses to these questions are A1 and A2. In the second stage, the input includes the medical record introduction and the Q\&A history from the first stage, and Q2 must be addressed. In the third stage, the input consists of the medical record introduction, the Q\&A history from the first two stages, and the patient's diagnosis and treatment process. In this stage, Q4 and Q5 must be answered sequentially.

\end{document}